\documentclass[11pt,twoside]{article}

\usepackage{fullpage}

\usepackage{amsmath,amssymb,fullpage,graphicx,xcolor,mathtools,nicefrac,comment}
\usepackage[shortlabels]{enumitem}

\usepackage{amssymb,amsmath,comment}
\usepackage[most]{tcolorbox}
\usepackage{subfig}

\usepackage{amsthm}
\makeatletter
\def\th@plain{%
  \thm@notefont{}% same as heading font
  \itshape % body font
}
\def\th@definition{%
  \thm@notefont{}% same as heading font
  \normalfont % body font
}
\makeatother

\usepackage[shortlabels]{enumitem}

\usepackage{thmtools,thm-restate}

\usepackage{hyperref}

\usepackage{amsmath,amssymb}
\usepackage{verbatim,float,url}
\usepackage{graphicx,psfrag}
\usepackage[numbers]{natbib}
\usepackage{xcolor}
\usepackage{microtype}
\usepackage{xparse}
\usepackage{xspace}
\usepackage{tikz,textcomp,thmtools,nameref,cleveref}
\usetikzlibrary{positioning}
\usepackage{tikz-qtree}

\usepackage{algpseudocode,algorithmicx,algorithm}

\usepackage[font=small]{caption}

\newtheorem{condition}{Condition}
\newtheorem{definition}{Definition}
\newtheorem{theorem}{Theorem}
\newtheorem{lemma}{Lemma}
\newtheorem{proposition}{Proposition}

\newcommand{\efflong}{Effective Planning Window} %our condition
\newcommand{\eff}{EPW} %acronym for our condition
\newcommand{\sdim}{d} %dimension of states
\newcommand{\tdim}{k} %dimension of theta
\newcommand{\pol}[3]{\pi_{#1}^{#2}(#3)}  % policy. 1 is state, 2 is action, 3 is parameter
\newcommand{\V}[2]{V_{\mathcal{M}}^{#1}(#2)}  % value. 1 is state, 2 is parameter
\newcommand{\tbound}{B} %bound on euclidean norm of theta
\newcommand{\plip}{\phi} %lipschitz constant of policy class
\newcommand{\failset}{\mathcal{F}} %set of failure states
\newcommand{\C}{C} %planning window
\newcommand{\trand}{\theta_{\mathrm{rand}}} %theta for random policy
\newcommand{\tvec}{\overline{\theta}} %vector of thetas
\newcommand{\argmax}{\mathop{\mathrm{argmax}}}
\newcommand{\argmin}{\mathop{\mathrm{argmin}}}
\newcommand{\Prob}{\mathbb{P}} %probability
\newcommand{\relu}{\mathrm{ReLU}}

\begin{document}

\begin{center}
{\bf{\LARGE{Sample Efficient Reinforcement Learning In \centerline{Continuous State Spaces:}\newline\centerline{A Perspective Beyond Linearity}}}}
%\cmcomment{this looks fine, another possibility:}

%{\bf{\LARGE{Optimality and Sub-optimality in Estimating Bivariate Isotonic Matrices with Unknown Permutations}}}

\vspace*{.2in}

{\large{
\begin{tabular}{cccc}
Dhruv Malik$^{\; \dagger}$ & Aldo Pacchiano$^{\; \ddagger}$ & Vishwak Srinivasan$^{\; \dagger}$ & Yuanzhi Li$^{\; \dagger}$
%p1$^\star$ & p2$^\dagger$ & p3$^{\dagger, \ddagger}$
\end{tabular}
}}
\vspace*{.2in}

%\begin{tabular}{cc}
%Machine Learning Department$^{\dagger}$ & Department of Electrical Engineering and Computer Sciences$^{\ddagger}$ \\
%Carnegie Mellon University & University of California, Berkeley \\
%Pittsburgh, PA 15213 & Berkeley, CA 94720
%\end{tabular}
\begin{tabular}{c}
Machine Learning Department, Carnegie Mellon University$^{\; \dagger}$ \\
Department of Electrical Engineering and Computer Sciences, UC Berkeley$^{\; \ddagger}$
\end{tabular}

\vspace*{.2in}

\today

\end{center}
\vspace*{.2in}
%%%%%%%%%%%%%%%%%%%%%%%%%%%%%%%%

\begin{abstract}
    Reinforcement learning (RL) is empirically successful in complex nonlinear Markov decision processes (MDPs) with continuous state spaces. By contrast, the majority of theoretical RL literature requires the MDP to satisfy some form of linear structure, in order to guarantee sample efficient RL. Such efforts typically assume the transition dynamics or value function of the MDP are described by linear functions of the state features. To resolve this discrepancy between theory and practice, we introduce the \efflong{} (\eff{}) condition, a structural condition on MDPs that makes \emph{no} linearity assumptions. We demonstrate that the \eff{} condition permits sample efficient RL, by providing an algorithm which provably solves MDPs satisfying this condition. Our algorithm requires minimal assumptions on the policy class, which can include multi-layer neural networks with nonlinear activation functions. Notably, the \eff{} condition is directly motivated by popular gaming benchmarks, and we show that many classic Atari games satisfy this condition. We additionally show the necessity of conditions like \eff{}, by demonstrating that simple MDPs with slight nonlinearities cannot be solved sample efficiently.
\end{abstract}

\section{Introduction}
\label{sec:intro}
Over the past decade, reinforcement learning (RL) has emerged as the dominant paradigm for sequential decision making in modern machine learning. During this time period, video games have served as popular means to benchmark the incremental improvement in state of the art RL. The Arcade Learning Environment (ALE), comprising a suite of classic Atari games, is an archetypical example of such a benchmark~\citep{bellemare13}. Agents trained by RL efficiently learn to surpass human level performance in such games~\citep{mnih13, mnih15, badia20}.

Motivated by these empirical accomplishments, there has been a major thrust to theoretically characterize the conditions which permit sample efficient RL. A significant line of work has greatly advanced our understanding of the tabular RL setting, where the number of states is finite and relatively small~\citep{simchowitz19, pananjady21}. Sample efficiency bounds in this setting scale with cardinality of the state space. However, in practice this cardinality is often large or infinite. For instance, many gaming applications of RL, ranging in complexity from Atari to Dota, all have continuous state spaces~\citep{berner19}. These scenarios are handled in the function approximation setting~\citep{du19, du20lowerbound}. Here, each state is associated with a known feature, and one desires a sample efficiency bound that scales with the dimensionality of the features (instead of the cardinality of the state space).

To understand when RL is sample efficient in continuous state spaces, theoreticians make certain assumptions on the features or the underlying Markov Decision Process (MDP). A prominent assumption, which has appeared in various forms, is that the problem satisfies some sort of \emph{linear} structure. For instance, in the well studied linear MDP, the transitions and rewards are described by linear functions of the features~\citep{yang19, jin20, yang20}. In particular, the transition probabilities at a state-action pair are defined by linear functions of the feature corresponding to that state-action pair. A weaker, but frequently occurring, form of this assumption is that value function of any policy is nearly linear~\citep{du20lowerbound, lattimore20}, or that the optimal value function is linear~\citep{du19, weisz21, weisz21lowerbound}. Such linear structure is amenable to theoretical analysis, since it permits analysts to leverage the vast literature on supervised and online learning.

To obtain a holistic understanding of RL, examining such linear structure is certainly important. Nevertheless, it is unclear whether the aforementioned linearity conditions actually hold in practical scenarios. We illustrate this via a very simple example. Consider an MDP where there is a set of $n$ states with the property that taking any action at one of these states leads to the same state. To cast this MDP in the aforementioned linear MDP setting, the dimensionality of the state features would have to scale linearly with $n$. This precludes the existence of algorithms that can solve this MDP with sample complexity independent of the cardinality of the state space.

Moreover, it has recently been shown both theoretically and empirically that the optimal value function and optimal policy can be very complex, even in ostensibly elementary continuous state space MDPs~\citep{dong20}. Since even powerful linear functions such as the Neural Tangent Kernel~\cite{jacot2018neural,li2018learning,als18dnn} are significantly worse in terms of representation power and robustness than nonlinear neural networks~\citep{allen-zhu19ntk, li20ntk,allen2020backward,allen2020feature}, it is unclear whether such weaker linear functions can be used to approximate the value function or the underlying policy well.

\begin{figure}[!tp]
\centering
\setlength{\fboxsep}{0mm}
\setlength{\fboxrule}{0.5mm}
\fcolorbox{black}{white}{\includegraphics[width=0.4\textwidth]{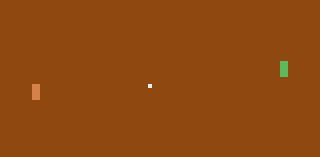}}
\caption{An image of the Atari Pong game. The green paddle must move up and down to hit the ball (the white dot) while playing against the opposing orange paddle.}
\label{fig:pong1}
\end{figure}

Even in simple RL gaming benchmarks, there is no evidence that the aforementioned linearity assumptions hold. Indeed, nonlinear neural networks are the predominant means to approximate policies and value functions, when solving these games in practice. For instance, consider the Pong game from the ALE benchmark, which is depicted in Figure~\ref{fig:pong1}. In this game, the agent must use its paddle to prevent the ball from crossing a boundary, while playing against a pseudorandom opposing paddle. Despite the simplicity of Pong, state of the art methods solve this game using neural networks~\citep{mnih13, mnih15, badia20}, and it is not apparent whether this game is linear in any sense.

This reveals a significant gap between the theory and practice of RL. In theory, one usually employs some sort of linearity assumption to ensure efficient RL. Yet, in practice, RL appears to succeed in domains which do not satisfy such linear structure. In an effort to resolve this discrepancy, we ask the following question:
\begin{center}
    \textbf{Which (\emph{non-linear}) structure is typical of popular RL domains, and how does this structure permit sample efficient RL?}
\end{center}
This question underlies the analysis of our paper. Towards answering this question, we make the following contributions:
\begin{itemize}
    \item We propose the \efflong{} (\eff{}) condition, a structural condition for MDPs which goes beyond linearity. Indeed, this condition is compatible with neural network policies, and MDPs satisfying \eff{} can have highly nonlinear (stochastic) transitions. Informally, this condition requires the agent to \emph{consistently plan $\C$ timesteps ahead}, for a value of $\C$ significantly smaller than the horizon length. We show that popular Atari benchmark games satisfy this condition.
    \item We provide a simple algorithm, which exploits the \eff{} condition to provably solve MDPs satisfying \eff{}. We prove the sample efficiency of our algorithm, and show that it requires a number of trajectories that is a lower order polynomial of the horizon length and other relevant problem dependent quantities.
    \item We argue that one must look beyond linear structure, and further motivate the study and necessity of conditions like \eff{}, by demonstrating that even slightly nonlinear MDPs cannot be solved sample efficiently.
\end{itemize}

\section{Related Work}
\textbf{Linear Function Approximation.} The majority of RL literature in the function approximation setting focuses on MDPs that satisfy some form of linear structure. A notable example is the linear MDP, where the transitions and rewards are described by linear functions of the state features~\citep{yang19, jin20, yang20}. A weaker form of this assumption is that the value function for each policy is nearly linear~\citep{du20lowerbound, lattimore20}, or that the optimal value function is linear~\citep{du19, weisz21, weisz21lowerbound}. We note that the algorithm of Weisz et al.~\citep{weisz21} requires a generative model, while we work in the standard episodic RL setting. As argued earlier, such linear assumptions are unlikely to hold true in practice. In our work, we eschew any sort of linearity assumption.

~\\ \textbf{Nonlinear Function Approximation.} Empirically, it is typical to use nonlinear function approximators such as neural networks~\citep{schulman15,levine16}. But from a theoretical perspective, the understanding of nonlinear function approximation is limited. There is prior work which studies the sample complexity of RL when using function approximation with nonlinear function classes~\citep{wen13, jiang17contextual, vanroy19, dong20generalfunc, du20, wang20generalfunc, jin21bellman, wang21}. However, these works often are restricted to MDPs with deterministic transitions~\citep{wen13, vanroy19, du20}. In this deterministic setting, an algorithm can repeatedly visit the same state and simply memorize an optimal path. By contrast, we focus on MDPs with \emph{stochastic} transitions, as is typical in many Atari games. Here, an algorithm generally cannot visit the same state more than once, and must generalize beyond the trajectories it samples to learn something global. Moreoever, the aforementioned analyses of nonlinear function approximation typically make some stringent assumption on the complexity of the function class~\citep{jiang17contextual, dong20generalfunc, wang20generalfunc, jin21bellman, wang21}. Such complexity measures either cannot or are not known to handle neural networks. By contrast, our results place minimal restrictions on the function class, and we can handle nonlinear multilayer neural networks. In a different line of work, Dai et al.~\citep{dai18} study RL with nonlinear function approximators and provide a convergent algorithm for this setting. However, they do not precisely relate the quality of the solution found by their algorithm to the approximation error of the function class.

~\\ \textbf{Linear Quadratic Regulator.} To characterize the sample complexity of RL in continuous state spaces, a different line of work investigates the linear quadratic regulator (LQR)~\citep{fazel18, dean19, malik20}. Here, the transition dynamics of the MDP are assumed to be noisy linear functions of the state and action, and the rewards are quadratic functions of the state and action. We remark that in this setting, the action space is continuous. By contrast, we exclusively study MDPs with finite action spaces, since these are most typical in the RL video game domains that motivate our paper.

\section{Problem Formulation}
\subsection{Problem Statement}
\label{sec:prob_statement}
\textbf{Notation \& Preliminaries.} We use the notation $[n]$ to denote $\{ 0, 1 \dots n-1 \}$ for any positive integer $n$. Recall that an undiscounted, finite horizon MDP $\mathcal{M} = (\mathcal{S}, \mathcal{A}, \mathcal{T}, R, H)$ is defined by a set of states $\mathcal{S}$, a set of actions $\mathcal{A}$, a transition function $\mathcal{T}$ which maps from state-action pairs to a probability density defined over states, a reward function $R$ which maps from state-action pairs to non-negative real numbers, and a finite planning horizon $H$. Throughout our paper, we assume that $\mathcal{S} \subseteq \mathbb{R}^{\sdim}$ and $\mathcal{A}$ is a finite set. Without loss of generality, we assume a single initial state $s_0$. For simplicity, we assume that $\mathcal{S}$ can be partitioned into $H$ different levels. This means that for each $s \in \mathcal{S}$ there exists a unique $h \in [H]$ such that it takes $h$ timesteps to arrive at $s$ from $s_0$. We say that such a state $s$ lies on level $h$, and denote $\mathcal{S}_h$ to be the set of states on level $h$. Note this assumption is without loss of generality, and our results apply to generic MDPs which cannot be partitioned into levels. This is because we can always make the final coordinate of each state encode the number of timesteps that elapsed to reach the state. Taking any action from level $H-1$ exits the game. The notation $\| x \|_2$ denotes the Euclidean norm of $x$.

A policy maps each state to a corresponding distribution over actions.
In practice, one typically uses a policy that is parameterized by parameters belonging to some set $\Theta \subseteq \mathbb{R}^{\tdim}$.
We study such policies, and use $\pol{}{}{\theta}$ to denote the policy induced by using parameter $\theta \in \Theta$. When discussing a policy which is not parameterized, we simply use $\pi$ to denote the policy.
We use $\pol{s}{a}{\theta}$ to denote the probability of taking action $a$ at state $s$ when using the policy $\pol{}{}{\theta}$.
We use $\pol{}{}{\Theta}$ to denote $\{ \pol{}{}{\theta} \text{ s.t. } \theta \in \Theta \}$, which is the set of feasible policies and defines our policy class.
Given a vector $\tvec \in \Theta^H$, we let $\pol{}{}{\tvec}$ denote the policy which executes $\pol{}{}{\tvec_h}$ at for any state lying on level $h \in [H]$, where $\tvec_h$ denotes the \(h^{th}\) entry of $\tvec$. The value of a policy $\pol{}{}{\theta}$ in a (stochastic) MDP $\mathcal{M}$ when initialized at state $s$ is denoted $\V{s}{\pol{}{}{\theta}}$. It is given by $\V{s}{\pol{}{}{\theta}} = \mathbb{E} \left[ \sum_{h=\text{level}(s)}^{H-1} R(s_h, a_h) \; \vert \; \pi(\theta) \right]$, where the expectation is over the trajectory $\{ (s_h, a_h) \}_{h=\text{level}(s)}^{H-1}$ conditioned on the fact that the first state in the trajectory is $s$.
Given an accuracy $\epsilon$ and failure probability tolerance $\delta$, the goal of RL is to find a policy $\pi$ which satisfies $\V{s_0}{\pi} \geq \max_{\pi'} \V{s_0}{\pi'} - \epsilon$ with probability at least $1 - \delta$.

~\\ \textbf{Query Model.} We adopt the standard episodic RL setup.
During each episode, an agent is allowed to interact with the MDP by starting from $s_0$, taking an action and to observe the next state and reward, and repeating.
The episode terminates after the agent takes $H$ actions, and the next episode starts at $s_0$.
The agent thus takes a single trajectory in each episode, and the total query complexity of the agent is measured by the total number of trajectories. Given a desired solution accuracy $\epsilon$ and failure probability tolerance $\delta$, we are interested in algorithms which can successfully solve an MDP using a number of trajectories that is at most polynomial in $H$, $\vert \mathcal{A} \vert$, $\sdim$, $\tdim$, $\frac{1}{\epsilon}$ and $\frac{1}{\delta}$.
If an algorithm provably accomplishes this, we call such an algorithm \emph{sample efficient} or \emph{tractable}.
Notably, such algorithms cannot depend on the (possibly uncountable) number of states.

~\\ Without any assumptions on the MDP, approximating an optimal policy is intractable.
To permit sample efficient RL, prior theoretical work has often assumed that MDP satisfies some form of linear structure.
For instance, the transition or value function might be described by a linear function of the states.
However, it is well documented that RL is empirically successful in highly nonlinear domains~\citep{schulman15, levine16}. We aim to bridge this gap between theory and practice. We now formally state the problem that we consider throughout our paper.
\begin{center}
    \emph{Our goal is to present nonlinear characteristic conditions which permit sample efficient RL, and argue that these conditions are satisfied in practice by popular RL domains.}
    %\emph{Our goal is to precisely characterize conditions which are sufficient for sample efficient RL, while focusing on conditions satisfied by popular RL domains, and to devise a sample efficient algorithm for these domains.}
\end{center}

\subsection{\efflong{} Condition}
\label{sec:eff}

We first state basic conditions that are satisfied by most RL problems encountered in practice. We will later refine these to obtain our \efflong{} (\eff{}) condition, and then show that \eff{} enables sample efficient RL.

Let us begin by observing that in practice, the policy class $\pol{}{}{\Theta}$ typically satisfies some mild regularity assumptions. We formalize this in the following condition.

\begin{condition}[Regular Policy Class]
\label{cond:regular-policy}
A policy class $\pol{}{}{\Theta}$ is said to be Regular when:
\begin{enumerate}[(a)]
    \item \textbf{Bounded Domain.} There exists $B > 0$ such that each $\theta \in \Theta$ satisfies $\| \theta \|_2 \leq \tbound$.
    \item \textbf{Lipschitz Continuous Policies.} There exists $\plip > 0$ such that for any $\theta, \theta' \in \Theta$ and any $(s, a) \in \mathcal{S} \times \mathcal{A}$, we have $\vert \pol{s}{a}{\theta} - \pol{s}{a}{\theta'} \vert \leq \plip \| \theta - \theta' \|_2$.
\end{enumerate}
\end{condition}

We stress that this is a very mild condition, and places minimal restrictions on $\pi(\Theta)$. Indeed, a policy parameterized by a multi-layer neural network with a nonlinear activation function satisfies this condition~\citep{fazlyab19}. Using this condition on the policy class, we now introduce the following Generic Game condition. As we will discuss in the sequel, many popular gaming RL benchmarks such as Atari games satisfy this condition.

\begin{condition}[Generic Game]
An MDP and Regular policy class pair $(\mathcal{M}, \pol{}{}{\Theta})$ form a Generic Game if:
\begin{enumerate}[(a)]
    \item \textbf{Failure States.} There is a set of failure states $\failset \subset \mathcal{S}$, and taking any action from a state in $\failset$ exits the game.
    \item \textbf{Complete Policy Class.} There exists some $\theta^\star \in \Theta$ such that executing $\pol{}{}{\theta^\star}$ from $s_0$ arrives at some state in $\mathcal{S}_{H-1} \setminus \failset$ almost surely\footnote{The Generic Game condition can also be defined in the case when this property of $\theta^\star$ holds true with probability exponentially large in $H$, as would occur when using a softmax policy class. Our results hold true when using this notion of a Generic Game. We focus on the almost sure case to avoid complicating notation.}. We define $\mathcal{S}^\star$ to be the set of all states $s \in \mathcal{S} \setminus \failset$ such that executing $\pol{}{}{\theta^\star}$ from $s$ reaches $\mathcal{S}_{H-1} \setminus \failset$ almost surely. If a state lies in $\mathcal{S}^\star$ we call it a safe state.
    \item \textbf{Binary Rewards.} For any state $s \in \mathcal{S}_{H-1} \setminus \failset$ and any $a \in \mathcal{A}$, $R(s,a) = 1$. For any other state $s$ and any $a \in \mathcal{A}$, $R(s, a) = 0$.
\end{enumerate}
\end{condition}

A few comments are in order.
Note that $\failset$ is essentially used to describe states where the agent has lost the game.
Also, observe that in Generic Games, an optimal policy is one that arrives at a non-failure state in level $H-1$ almost surely.
Hence $\pol{}{}{\theta^\star}$ is indeed an optimal policy.

\begin{figure}[!tp]
    \centering
    \setlength{\fboxsep}{0mm}
    \setlength{\fboxrule}{0.5mm}
    \fcolorbox{black}{white}{\includegraphics[width=0.4\textwidth]{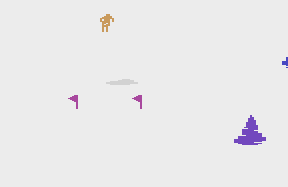}}
    \caption{An image of the Atari Skiing game. The skier must move through flagged checkpoints.}
    \label{fig:skiing1}
\end{figure}

Let us now describe how popular Atari games can be cast as Generic Games.
Recall the famous Pong game depicted in Figure~\ref{fig:pong1}, which is a part of the ALE benchmark~\citep{bellemare13}.
In this game, an RL agent must learn to move the paddle up and down to hit the ball and prevent it from crossing its boundary.
Note that in the context of RL, this is a single player game, since the opposing paddle hits the ball back according to a pre-specified stochastic decision rule (which is not trained). The agent loses the game if the ball crosses its own boundary, and wins the game if it hits the ball past the opposing paddle. Another game in the ALE benchmark is the Skiing game, depicted in Figure~\ref{fig:skiing1}. Here, the agent must move the skier through a series of randomly appearing flagged checkpoints, which appear frequently over a long time horizon.
The skier receives a penalty each time it misses a checkpoint.

We claim that an Atari game like Pong or Skiing, together with a neural network policy class, satisfy the Generic Game condition.
The first two conditions of Generic Games are easy to verify.
Note that the states in Pong (resp. Skiing) are images, so $\failset$ includes any state where the ball has crossed the agent's boundary (resp. where the skier has missed a prior checkpoint), since this corresponds to the agent failing to complete the game.
It is well known that Atari can be solved using a neural network policy~\citep{mnih15}, so a policy class parameterized by neural networks is indeed complete.

To ensure that Pong and Skiing satisfy the third condition, we need to design an appropriate binary reward function. For Pong, this is handled by redefining $\failset$ to include any state $s \in \mathcal{S}_{H-1}$ where the ball has not crossed the opposing paddle. Similarly for Skiing, this is done by ensuring $\failset$ includes any state where the skier has already missed a checkpoint. Then one can simply assign a reward of $1$ to any state in $\mathcal{S}_{H-1} \setminus \failset$, and $0$ to all other states, as required by the Generic Game condition. Hence, playing optimally in this Generic Game framework ensures that the ball has moved past the opposing paddle, or that the skier has made all checkpoints, corresponding to winning the game.

The aforementioned reward design is an example of reward shaping, which is unavoidable in RL and ubiquitous in practice~\citep{hadfield-menell17}.
Nevertheless, we stress that the reward function we described above is very similar to the reward function that practitioners already use.
Concretely, in Pong one typically assigns a reward of $1$ if the ball has moved past the opposing paddle, a reward of $-1$ if the ball has moved past the agent's paddle, and a reward of $0$ otherwise~\citep{bellemare13}.
Similarly, in Skiing, the skier receives reward at the end of the game, in proportion to the number of checkpoints it has cleared~\citep{badia20}. Our reward function thus requires no more effort to design than the reward functions already in use, since they require the same information, and these are identical in spirit.

Beyond Pong and Skiing, other Atari games (and other similarly themed video games) can be cast in the Generic Game framework. In Appendix~\ref{app:games}, we describe this reduction for the Atari games Tennis and Journey Escape, and also for the more complex RL gaming benchmark CoinRun~\cite{cobbe19}.

Let us make one more remark about the binary reward structure of Generic Games. There are many games which naturally have a set of goal states, and these immediately can be described as Generic Games. Examples include Pong, Tennis and CoinRun (latter two are described in Appendix~\ref{app:games}). More generally, however, EPW applies to many games which do not naturally have a binary reward structure. As we discussed, Skiing can be cast in the EPW framework with binary rewards, by ensuring that $\mathcal{F}$ includes any state where the skier has missed a checkpoint. However, in certain scenarios, one may be satisfied with only collecting a large fraction of the checkpoints in Skiiing, or more generally, obtaining a large (but not perfect) score in games where one continually collects small rewards. We emphasize that such scenarios can be cast in our Generic Game framework. For instance, in Skiing if checkpoints arrive roughly every $x$ timesteps, and we desire to give a reward of $1$ for each checkpoint collected, then we can design $\mathcal{F}$ to include any state at timestep $t$ where the agent has not collected $\Omega(t/x)$ reward thus far. Similar reductions apply to other games where one continually needs to collect a small amount of reward at regular intervals.

Does the Generic Game condition permit sample efficient RL? Unfortunately, there exist Generic Games where the MDP is only slightly nonlinear, but even approximating an optimal policy sample efficiently is impossible. We later show this formally in Proposition~\ref{prop:lower_nonlinear}. So we must further restrict this class of games. In order to refine our notion of a Generic Game, we first state a useful definition.

\begin{definition}[$\boldsymbol{x}$-Ancestor]
Given a Generic Game $(\mathcal{M}, \pol{}{}{\Theta})$, consider any $h \in [H]$ and any state $s' \in \mathcal{S}_h$. A state $s \in \mathcal{S}$ is an $x$-ancestor of $s'$, if $s \in \mathcal{S}_{\max \{ 0, h - x \} }$ and there exists some $\theta \in \Theta$ such that following $\pol{}{}{\theta}$ from $s$ will reach $s'$ with nonzero probability.
\end{definition}

We are now in a position to formally state our \efflong{} (\eff{}) condition, which refines our notion of Generic Games. For the statement of the condition, recall our notion of $\mathcal{S}^\star$, which was defined in the Generic Game condition.

\begin{condition}[\efflong{}]
A Generic Game $(\mathcal{M}, \pol{}{}{\Theta})$ satisfies the \efflong{} condition with parameter $\C$ if there exists $\C \in [H]$ such that the following holds. Consider any $s' \in \mathcal{S} \setminus \failset$. If $s$ is a $\C$-ancestor of $s'$, then $s \in \mathcal{S}^\star$.
\end{condition}

Before examining RL benchmark games in the context of this condition, a few comments about the condition itself are in order.
The quantity $\C$ ensures that any $\C$-ancestor of a non-failure state is a safe state.
So if an agent is at timestep $t$ and the game is not over, then at timestep $t - \C$ it was in a state from where it could have achieved the highest reward possible in the MDP (if it took the correct sequence of actions).
For the purposes of RL, this effectively means that at each timestep, the agent must consistently plan over the next $\C$ timesteps instead of the entire horizon length $H$. Thus, when $\C$ is small or a constant, then it is reasonable to believe that sample efficient RL is possible.

Of course, any Generic Game satisfies the \eff{} condition for a choice of $\C = H-1$.
However, many popular RL benchmark games satisfy the \eff{} property with a value of $\C$ that is much smaller than $H$. Informally, the $\C$ quantity is the amount of time required by the agent to successfully \emph{react} to scenarios in the game (without losing). Let us understand this more deeply in the Pong and Skiing games.

\begin{figure*}[!tp]
\captionsetup[subfigure]{labelformat=empty}
\setlength{\fboxrule}{1mm}
\setlength{\fboxsep}{0mm}
\centering
\subfloat[\(s_{t''} \in \mathcal{S}^{\star}\)]{\fcolorbox{green}{white}{\includegraphics[width=0.175\textwidth]{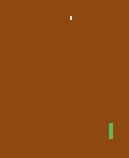}}}
\hfill
\subfloat[\( s_{t-\C} \in \mathcal{S}^{\star}\)]{\fcolorbox{green}{white}{\includegraphics[width=0.175\textwidth]{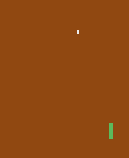}}}
\hfill
\subfloat[\( s_{t'} \in \mathcal{S} \setminus \left( \mathcal{S}^{\star} \cup \failset \right)\)]{\fcolorbox{yellow}{white}{\includegraphics[width=0.175\textwidth]{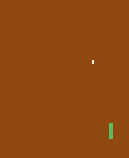}}}
\hfill
\subfloat[\( s_t \in \mathcal{S} \setminus \left( \mathcal{S}^{\star} \cup \failset \right) \)]{\fcolorbox{yellow}{white}{\includegraphics[width=0.175\textwidth]{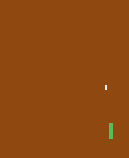}}}
\hfill
\subfloat[\( s_{t+1} \in \failset\)]{
\fcolorbox{red}{white}{\includegraphics[width=0.175\textwidth]{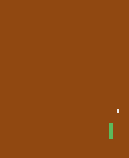}}}

\caption{Five states from the Pong game. Here we let $t'' < t - \C < t' < t$, and the ball is progressively moving towards the lower right corner. At timesteps $t', t$, the paddle has not lost the game. However, it does not have enough time to react and reach the ball in time. At timestep $t+1$ the game is over. At timesteps $t'', t-\C$, the paddle has enough time to react and reach the ball.}
\label{fig:pong-in-motion}
\end{figure*}

In Pong, after the opposing paddle hits the ball, the agent must react to the trajectory of the ball and adjust its position accordingly to hit it. If it takes too long to react before it starts adjusting its position, then it will be unable to reach the ball in time. We depict this in Figure~\ref{fig:pong-in-motion}. More formally, assume that at timestep $t$ the paddle has not lost the game and the ball is moving towards its boundary. At timestep $t$, the ball may be too close to the boundary, and so the agent will not not have enough time to move its paddle fast enough in order to reach the ball in time. However, at timestep $t-\C$ the ball is further away from the boundary, so the agent has enough time to move its paddle appropriately in order to react, reach the ball and hit it back. So at timestep $t-\C$ the agent lies in a safe state in $\mathcal{S}^\star$, since it has enough time to adjust its paddle and hit the ball back, and hence play optimally. Notably, if we let $\C'$ be the number of timesteps it takes for the ball to traverse from one end of the board to the other, then $\C \leq \C'$. Hence, when $H$ is large and the agent needs to control the paddle for many rounds, then $\C$ is a constant independent of $H$.

\begin{figure*}[!tp]
\captionsetup[subfigure]{labelformat=empty}
\setlength{\fboxrule}{1mm}
\setlength{\fboxsep}{0mm}
\centering
\subfloat[\(s_{t''} \in \mathcal{S}^{\star}\)]{\fcolorbox{green}{white}{\includegraphics[width=0.175\textwidth]{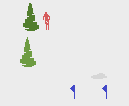}}}
\hfill
\subfloat[\( s_{t-\C} \in \mathcal{S}^{\star}\)]{\fcolorbox{green}{white}{\includegraphics[width=0.175\textwidth]{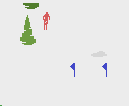}}}
\hfill
\subfloat[\( s_{t'} \in \mathcal{S} \setminus \left( \mathcal{S}^{\star} \cup \failset \right)\)]{\fcolorbox{yellow}{white}{\includegraphics[width=0.175\textwidth]{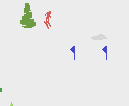}}}
\hfill
\subfloat[\( s_t \in \mathcal{S} \setminus \left( \mathcal{S}^{\star} \cup \failset \right) \)]{\fcolorbox{yellow}{white}{\includegraphics[width=0.175\textwidth]{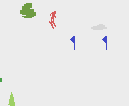}}}
\hfill
\subfloat[\( s_{t+1} \in \failset\)]{
\fcolorbox{red}{white}{\includegraphics[width=0.175\textwidth]{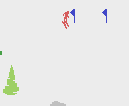}}}

\caption{Five states from the Skiing game. Here we let $t'' < t - \C < t' < t$, and the skier is progressively moving towards the checkpoint. At timesteps $t', t$, the skier has not lost. However, it does not have enough time to react and adjust its position to reach the checkpoint in time. At timestep $t+1$ the game is over. At timesteps $t'', t-\C$, the skier has enough time to react and reach the checkpoint.}
\label{fig:skiing-in-motion}
\end{figure*}

Similarly, in the Skiing game, the skier must react to the location of the oncoming checkpoint, and adjust its position accordingly. Formally, assume at timestep $t$ a checkpoint is oncoming. In such a scenario, as depicted in Figure~\ref{fig:skiing-in-motion}, the skier might be too far from the checkpoint in order to actually clear it (even if it moves directly towards the checkpoint). However, at timestep $t - \C$ the skier has enough time to adjust its position in order to clear the checkpoint. Hence at timestep $t-\C$, the skier is in a safe state in $\mathcal{S}^\star$ since it can play optimally from this state. Again, if we let $\C'$ be the number of timesteps it takes for a skier to move from the left edge of the screen to the right edge, then $C \leq C'$. Hence, when $H$ is large and there are many checkpoints to be cleared, then $\C$ is a constant independent of $H$, as previously observed in Pong.

Beyond Pong and Skiing, other Atari games satisfy the \eff{} condition, with a constant value of $\C$. We demonstrate this for the Atari games Tennis \& Journey Escape in Appendix~\ref{app:games}. In Appendix~\ref{app:games} we additionally show that more complex games, such as CoinRun~\citep{cobbe19}, also satisfy \eff{} with a small value of $\C$. We stress that EPW is orthogonal to linearity. Indeed, there are MDPs satisfying linearity but not EPW, and vice versa. We conclude this section by highlighting two important aspects of the \eff{} condition.

~\\ \textbf{The Magnitude Of $\boldsymbol{\C}$.} We treat $\C$ as a constant that is independent of and much smaller than $H$. This is certainly reasonable given our above discussion. So an algorithm incurring $\mathcal{O}(\vert \mathcal{A} \vert^\C)$ sample complexity is efficient. Furthermore, as we discuss later, there exist \eff{} games where $\Omega(\vert \mathcal{A} \vert^\C)$ sample complexity is necessary to solve the game.

~\\ \textbf{The Challenge Of Solving \eff{} Games.} We note that a deterministic \eff{} game is straightforward to solve, since an agent can just try each of the $\vert \mathcal{A} \vert^\C$ trajectories when it is at level $h$, to discover which trajectories do not lead to $\failset$. In such a case, an agent can simply memorize a good path. However, when transitions are stochastic (as in Atari), the agent cannot simply try each trajectory to memorize one that does not lead to $\failset$. This is because in general stochastic MDPs, a finite sample algorithm might only visit any given state at most once. Instead, the algorithm must \emph{learn} and generalize beyond the trajectories it samples, to learn something global about the MDP. Furthermore, we emphasize that stochastic \eff{} games \emph{cannot} be solved as simply as just splitting the horizon $H$ into $H/\C$ distinct planning windows, and then solving these planning problems independently of each other. Instead, the key difficulty is that the agent must \emph{consistently} plan $\C$ timesteps ahead. By this, we mean that just because an agent has arrived at a non-failure state at time $t$, does not imply that at time $t+1$ it is guaranteed to avoid $\mathcal{F}$. Indeed, if we execute a policy and the resulting trajectory ends in a failure state after $t$ timesteps, then it is unclear at which of the prior timesteps $\{ t-\C \dots t-1\}$ that we took an incorrect action. And we cannot rollback to timestep $t-\C$ and rerun the same trajectory to discover when we made a mistake. This complicates the design of efficient algorithms for this setting.

\section{Main Results}
We now turn to our main results.
Before diving into the details, let us provide a brief overview for the results.
Our main contribution is an algorithm which sample efficiently solves games satisfying the \eff{} condition.
We prove the efficiency of this algorithm and discuss the role of the \eff{} condition in permitting sample efficient RL.
We then further motivate the study of conditions like \eff{}, as opposed to the study of linear structure, by proving a lower bound which shows that Generic Games with even slight nonlinearities cannot be solved sample efficiently. Hence, we must look beyond linear structure to characterize when sample efficient RL is possible in realistic domains. With this outline in mind, let us now formally present our main results.

Our primary contribution is an algorithm which exploits the \eff{} condition to sample efficiently find an optimal policy.
Our algorithm is defined in Algorithm~\ref{alg:main}.
For this method, recall our notation that for a vector $\tvec \in \Theta^H$, we let $\pol{}{}{\tvec}$ denote the policy which executes $\pol{}{}{\tvec_h}$ at each $h \in [H]$, where $\tvec_h$ denotes the \(h^{th}\) entry of $\tvec$.
We make the basic assumption that there exists $\trand \in \Theta$ such that $\pol{}{}{\trand}$ maps each state to the uniform distribution over $\mathcal{A}$.

\begin{algorithm}[t]
\caption{}
\label{alg:main}
\begin{algorithmic}[1]
\State Inputs: MDP $\mathcal{M}$, policy class $\pol{}{}{\Theta}$, sample size $n$
\State Initialize $\tvec(0) = [\trand, \trand \dots \trand] \in \Theta^{H-1}$
\For{$t \in \{ 0, 1 \dots H-2 \}$}
\State Sample $n$ trajectories $\{ \tau_i \}_{i=1}^n \sim \pol{}{}{\tvec(t)}$, where each trajectory $\tau_i = \{ (s_{i, h}, a_{i, h}) \}_{h=0}^{H-1}$
\State Define the loss function $\widehat{L}_t: \Theta \to \mathbb{R}$ as
\begin{equation}
\label{eq:emp_loss_defn}
\widehat{L}_t(\theta) = \vert \mathcal{A} \vert^{\C + 1} \cdot \left[ \frac{1}{n} \sum_{i=1}^n \left( \mathbb{I}_{s_{i, t+C+1} \in \failset} \prod_{j=0}^C \pol{s_{i, t+j}}{a_{i, t+j}}{\theta} \right) \right]
\end{equation}
\State Minimize $\widehat{L}_t$ over $\Theta$ to obtain
\begin{equation}
\label{eq:emp_loss_minimizer}
    \widehat{\theta}_t \in \argmin_{\theta \in \Theta} \widehat{L}_t(\theta)
\end{equation}
%\STATE Update \(t^{th}\) entry of $\tvec$ as $\theta_t$
\State Define \(\tvec(t+1) = [\widehat{\theta}_0, \widehat{\theta}_1 \ldots, \widehat{\theta}_{t-1}, \widehat{\theta}_t, \trand, \ldots, \trand]\)
\EndFor
\State \Return $\tvec(H-1)$
\end{algorithmic}
\end{algorithm}

\begin{comment}

\begin{algorithm}
\caption{GenRL}
\label{alg:main}
\begin{algorithmic}[1]
\State \text{Inputs: horizon length $H$, distribution $\mathcal{D}$, sample size $n$}
%\State Initialize $\mathcal{S}_\pi$ and $\mathcal{A}_\pi$ each as the empty set
%\State Initialize $\pi$ as the empty function from $\mathcal{S}_\pi$ to $\mathcal{A}_\pi$
\State Initialize $\pi$ as the empty function
\For{$t \in \{0, 1 \dots H-1 \}$}
\State Sample $\{ M_i \}_{i=1}^n \overset{i.i.d}{\sim} \mathcal{D}$
\For{$a \in \ActSet$}

\State Let $s'$ be the child state of $s_t$ when taking action $a$
\State Query $\oracle$ to obtain $\oracle(s', M_i) = \approxV{s'}{M_i}$
\State Store $Q_{i, a} = \Rew_{M_i}(s_t, a) + \approxV{s'}{M_i}$
\EndFor

\State Take action $a_t = \argmax_{a' \in \ActSet} \{ \frac{1}{n} \sum_{i=1}^n Q_{i, a'} \}$ and arrive at state $s_{t+1}$
%\State Update $\mathcal{S}_\pi = \mathcal{S}_\pi \cup \{ s_t \}$ and $\mathcal{A}_\pi = \mathcal{A}_\pi \cup \{ a_t \}$
\State Define $\pi(s_t) = a_t$
\EndFor
\State \Return $\pi$
\end{algorithmic}
\end{algorithm}

\end{comment}

In our main result Theorem~\ref{thm:main}, we bound the sample complexity of Algorithm~\ref{alg:main} and demonstrate that it efficiently finds a near optimal policy for any \eff{} game.
Before stating the theorem, let us discuss Algorithm~\ref{alg:main} and provide some intuition for this method.
Recall that in Generic Games, a near optimal policy avoids failure states with high probability.
The method initializes $\tvec(0)$ to be the parameter vector which induces a uniformly random policy regardless of the state (Line 2).
It then incrementally updates $\tvec(t)$, in a fashion that ensures $\pol{}{}{\tvec(t)}$ avoids failure states with high probability for each $t' \leq t$ (Line 7). Ultimately it returns the parameter vector $\tvec(H-1)$ (Line 9).
More concretely, at each timestep $t$ in the inner loop, the algorithm samples $n$ trajectories from the policy $\pol{}{}{\tvec(t)}$ that it has constructed thus far (Line 4).
Via these sampled trajectories it defines the empirical loss $\widehat{L}_t$, as shown in Eq.~\eqref{eq:emp_loss_defn}. Intuitively, $\widehat{L}_t$ penalizes those parameters $\theta$ such that executing $\pol{}{}{\theta}$ over timesteps $\{ t, t+1 \dots t+C \}$ arrives at a failure state with high probability.
This intuition suggests that a minimizer $\widehat{\theta}_t$ of $\widehat{L}_t$, as defined in Eq.~\eqref{eq:emp_loss_minimizer}, will assign low probability to trajectories ending in a failure state when playing $\widehat{\theta}_t$.
The quantity $\theta_t$ then defines the $t+1^{th}$ entry of $\tvec(t+1)$.
Note that $\tvec(t+1)$ agrees with $\tvec(t)$ in its first $t$ entries.

The form of the loss $\widehat{L}_t$ suggests that a pracitioner needs to know the exact value of $\C$ to use Algorithm~\ref{alg:main}. This may be a stringent requirement in practice. We emphasize that any upper bound $\C' \geq \C$ can be used in place of $\C$ in the definition of $\widehat{L}_t$.
As discussed in Section~\ref{sec:eff}, such an upper bound is easy to find in gaming domains where we expect the \eff{} condition to hold. For example, in Pong we can play the game manually (in OpenAI Gym), and observe the number of timesteps it takes the paddle to traverse its side. This yields $C \leq 15$, while $200 \leq H$. As to how this affects the sample complexity of the method, one can simply substitute $\C'$ for $\C$ in the bound provided in our main result Theorem~\ref{thm:main}.

Before formally presenting Theorem~\ref{thm:main}, a remark on the computational requirements of Algorithm~\ref{alg:main} is imperative.
The method requires oracle access to a minimizer $\widehat{\theta}_t$ of the loss $\widehat{L}_t$, which in turn is defined by the policy class $\pol{}{}{\Theta}$.
In our paper, we impose minimal assumptions on $\pol{}{}{\Theta}$. Our motivation for this choice is that in practice, it is most common to parameterize a policy via a multi-layer neural network with nonlinear activation function.
Beyond our extremely mild Regularity condition, is unclear which (if any) desirable properties such a policy class satisfies. Hence, for a worst case Regular policy class $\pol{}{}{\Theta}$, obtaining even an approximation of $\widehat{\theta}_t$ could be extremely computationally intractable.
Nevertheless, we stress that in both theory and practice, stochastic gradient descent and its variants have been shown to efficiently find global minima of loss functions parameterized by neural networks~\citep{li18, allen-zhu19, du19gdnn}. Furthermore, as we show in our proofs, the function $\widehat{L}_t$ is Lipschitz continuous with a tolerable Lipschitz constant. Hence by Rademacher's Theorem, $\widehat{L}_t$ is differentiable almost everywhere, and it is reasonable to minimize $\widehat{L}_t$ via the stochastic gradient type methods that are popular for minimizing complex neural network losses. We also remark that it is fairly common in the RL literature to assume access to a computational oracle, when studying sample complexity~\citep{agarwal14, du19decoding, agarwal20flambe, misra20}.

We now formally state Theorem~\ref{thm:main}, our main result. This theorem bounds the sample complexity required by Algorithm~\ref{alg:main} to find a near optimal policy for any \eff{} game.

\begin{theorem}
\label{thm:main}
Fix error tolerance $\epsilon > 0$ and failure probability tolerance $\delta > 0$. Given any $(\mathcal{M}, \pi(\Theta))$ satisfying the \eff{} condition, and sample size

\begin{equation*}
    n = \frac{4H^{2} \vert \mathcal{A} \vert^{2\C + 2}}{\epsilon^{2}} \left( \log\left(\frac{2H}{\delta}\right) + \tdim \log \left(1 + \frac{32 H \vert \mathcal{A} \vert^{\C + 1} \C \plip \tbound}{\epsilon} \right) \right),
\end{equation*}
Algorithm~\ref{alg:main} outputs $\tvec$ satisfying
$$
\V{s_0}{\pol{}{}{\tvec}} \geq \V{s_0}{\pol{}{}{\theta^\star}} - \epsilon
$$
with probability at least $1 - \delta$.
\end{theorem}

A formal proof for the theorem is presented in Appendix~\ref{app:upper_bound}. A few comments are in order.
Note that Algorithm~\ref{alg:main} samples $n$ trajectories at each timestep in its inner loop.
Recall from the definition of our query model in Section~\ref{sec:prob_statement}, that we measure the total sample complexity by the total number of trajectories sampled.
Hence, the total sample complexity of Algorithm~\ref{alg:main} scales as $\mathcal{O}(\frac{H^3 \vert \mathcal{A} \vert^{2\C+2} \tdim}{\epsilon^2})$, where we have discarded logarithmic factors. Observe that the total sample complexity depends only logarithmically on the failure probability tolerance $\delta$, the bound $\tbound$ on the Euclidean norm of $\Theta$ and the Lipschitz constant $\plip$ of the policy.
It is also worth noting that the sample complexity has \emph{no} explicit dependence on the dimension $\sdim$ of states. However, it does depend linearly on the dimension $\tdim$ of the policy parameter space $\Theta$, and of course in general $\tdim$ will scale with $\sdim$. We note that in practical RL, one typically employs a shallow neural net, with only two or three layers. $\tdim$ is relatively small in this regime, in contrast to NLP or vision tasks where models are much larger.

Observe that the sample complexity bound in Theorem~\ref{thm:main} has an exponential dependence on $\C$. Recall that in our framework, as motivated at the end of Section~\ref{sec:eff}, $\C$ is a constant, so our algorithm is indeed efficient. Nevertheless, we remark that as a direct corollary of the work of Du et al.~\citep{du20lowerbound}, this exponential dependence on $\C$ cannot be improved by a better algorithm or sharper analysis.

More generally, in the context of existing literature, we provide some intuition for why the \eff{} condition permits efficient learning. A major issue that hinders sample efficient RL, in both theory and practice, is that an agent must plan over the entire horizon $H$. Roughly speaking, at each timestep the agent can choose any of $\vert \mathcal{A} \vert$ actions, so the total sample complexity required to plan over $H$ timesteps scales as $\Omega(\vert \mathcal{A} \vert^H)$. This is a recurrent theme in the RL literature, and various prior works have shown that even when the MDP has some non-trivial structure, in the worst case such a scaling is unavoidable~\citep{du20lowerbound, malik21}. Assuming that the MDP satisfies significant \emph{linear} structure, is one way to avoid this difficulty. By contrast, we are able to avoid this difficulty while making \emph{no} linearity assumptions. Instead, the \eff{} condition guarantees that an agent needs only to plan over $\C$ timesteps. Hence we exchange the worst case $\Omega(\vert \mathcal{A} \vert^H)$ scaling for the benign $\mathcal{O}(\vert \mathcal{A} \vert^{2\C + 2})$ scaling.

We believe that the \eff{} condition (or other conditions that are similar in spirit) is the \emph{correct} condition for characterizing when sample efficient RL is possible, at least in RL domains like video games.
By contrast, the linearity assumptions which prominently appear in prior literature, in addition to lacking clear empirical justification, are quite brittle.
To demonstrate this, we leverage prior work to show the existence of Generic Games which have only slight nonlinearities, yet cannot be solved sample efficiently.
Before we state this lower bound, we recall two standard definitions.

\begin{definition}[Optimal Value Function]
The optimal value function $V_{\mathcal{M}}^\star: \mathcal{S} \to \mathbb{R}$ of an MDP $\mathcal{M}$ is defined as $V_{\mathcal{M}}^\star(s) = \V{s}{\pi^\star}$, where $\pi^\star$ is an optimal policy of $\mathcal{M}$.
\end{definition}

\begin{definition}[Softmax Linear Policy]
For an MDP $\mathcal{M}$, a softmax linear policy $\pol{}{}{\boldsymbol{\theta}}$ is parameterized by $\boldsymbol{\theta} \in \mathbb{R}^{\vert \mathcal{A} \vert \times d }$. Letting $\boldsymbol{\theta}_i$ denote the $i^{th}$ row of $\boldsymbol{\theta}$, the policy $\pol{}{}{\boldsymbol{\theta}}$ satisfies
$$
\pol{s}{a_i}{\boldsymbol{\theta}} = \frac{\exp(s^T \boldsymbol{\theta}_i) }{\sum_{a_j \in \mathcal{A}} \exp(s^T \boldsymbol{\theta}_j)}.
$$
\end{definition}

Briefly, the optimal value function takes as input a state and outputs the optimal value that one can achieve from that state. A softmax linear policy is parameterized by a matrix, whose rows are in correspondence with actions. Given a state, the softmax linear policy outputs a probability distribution over actions, where the probabilities are exponentially weighted linear functions of the state.

We now demonstrate the existence of Generic Games, which have only slight nonlinearities, where even approximating an optimal policy in a sample efficient manner is impossible.
This result is heavily inspired by the recent work of Du et al.~\citep{du20lowerbound}, and we claim no technical novelty. Rather, the purpose of this result in our setting, is to further motivate the importance of studying conditions such as \eff{}, instead of assuming that the MDP has linear structure.

\begin{proposition}
\label{prop:lower_nonlinear}
There exists a Generic Game $(\mathcal{M}, \pol{}{}{\Theta})$, where $\sdim$, $\tdim$, $\plip$, $\tbound$ are all at most polynomial in $H$ and $\vert \mathcal{A} \vert$, and $\pol{}{}{\Theta}$ is the class of softmax linear policies, such that the following holds.
There exists an unknown neural network $f: \mathbb{R}^{\sdim} \to \mathbb{R}$, where $f$ is a linear combination of two $\relu$ neurons, such that $V^\star_{\mathcal{M}}(s) = f(s)$ for all $s \in \mathcal{S}$. Yet, any algorithm requires $\Omega(\min \{ \vert \mathcal{A} \vert^H, 2^\sdim \})$ trajectories to find, with probability at least $\nicefrac{3}{4}$, a policy $\pi$ satisfying
$$
\V{s_0}{\pi} \geq \V{s_0}{\pol{}{}{\theta^\star}} - \nicefrac{1}{4}.
$$
\end{proposition}

We stress that the essence of this result follows from Du et al.~\citep{du20lowerbound}, and we only make small modifications to their proof to fit it in our Generic Game setting. We nevertheless provide a proof sketch in Appendix~\ref{app:lower_bound}. The result shows the existence of MDPs for which a softmax linear policy is optimal, and where the optimal value function can be expressed as a neural network with only two $\relu$ neurons.
 Despite only this slight nonlinearity, sample efficient RL is impossible. Notice that in the statement of this result, there is no dependency of $\pi$ on $\theta$.
 This is because we do not restrict the algorithm to only search over policies lying in $\pol{}{}{\Theta}$. In particular, the policy $\pi$ mentioned in the result can be \emph{arbitrary}, and does not have to lie in $\pol{}{}{\Theta}$.
 
 Proposition~\ref{prop:lower_nonlinear} demonstrates that if the Generic Game is even slightly nonlinear, as one would expect in practice, sample efficient RL is impossible.
 So we must look beyond linearity to obtain a realistic characterization of when sample efficient RL is possible.
 Our \eff{} condition, which makes no linearity assumptions, is one example of this.

\section{Discussion}
In this paper, we studied structural conditions which permit sample efficient RL in continuous state spaces, with a focus on conditions that are typical in popular RL domains such as Atari games. We introduced the \eff{} condition, which in contrast to prior work, makes no linearity assumptions about the MDP structure. We provided an algorithm which provably solves MDPs satisfying \eff{}. We analyzed the sample complexity of this algorithm, and showed it requires a number of trajectories that is a lower order polynomial of the horizon length and other relevant problem dependent quantities. We also showed that MDPs which have very slight nonlinearities (but do not satisfy \eff{}) cannot be solved sample efficiently. Our analysis thus highlights the important need to look beyond linear structure, in order to establish the sample efficiency of RL in popular domains.

A number of open questions remain. First, while our \eff{} condition is directly motivated by RL gaming domains such as Atari, it is unclear whether \eff{} is satisfied by other RL application domains such as robotics. A natural direction for future work is to study these domains more closely, and identify structure that permits sample efficient RL in such domains. Second, recall that our algorithm requires access to a particular computational oracle. As discussed, we made this computational abstraction since we placed minimal restrictions on the policy class, so in the worst case obtaining such an oracle could be intractable. Nevertheless, we suspect that when using a neural network policy class with an appropriate architecture, one could approximate this oracle efficiently. It would be interesting to precisely characterize when this is possible. Third, it would be interesting to see whether a variant of our theoretically justified algorithm can be deployed in practice. Using our theoretical insight to design a pragmatic method, with strong empirical performance, is an important direction for future work.

\subsection*{Acknowledgements}
This material is based upon work supported by the National Science Foundation Graduate Research Fellowship Program under Grant No. DGE1745016. Any opinions, findings, and conclusions or recommendations expressed in this material are those of the authors and do not necessarily reflect the views of the National Science Foundation.

\appendix
\section{Other Games Satisfying \eff{}}
\label{app:games}
In this section, we shall verify that several other games besides Pong and Skiing satisfy the \eff{} condition. In Appendix~\ref{app:atari} we will verify this for the Atari games Tennis and Journey Escape. In Appendix~\ref{app:non_atari} we will verify this for the RL gaming benchmark CoinRun~\citep{cobbe19}, which is more complex than Atari.

\subsection{Atari Games}
\label{app:atari}

\begin{figure}[H]
\centering
\setlength{\fboxsep}{0mm}
\setlength{\fboxrule}{0.5mm}
\fcolorbox{black}{white}{\includegraphics[width=0.4\textwidth]{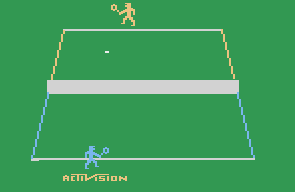}}
\caption{An image of the Atari Tennis game. The yellow player must move to hit the ball (the white dot) while playing against the opposing blue player.}
\label{fig:tennis1}
\end{figure}

\textbf{Tennis.} This game is very similar to Pong, and is depicted in Figure~\ref{fig:tennis1}. Here the agent controls the tennis player depicted at the top of the screen, who must hit the ball and prevent it from crossing its boundary. It plays against a player which hits the ball back according to a pre-specified stochastic decision rule (which is not trained). The agent loses if the ball crosses its own boundary, and wins if it hits the ball past the opponent's boundary.

The first two conditions of Generic Games are easy to verify. Note that the states in Tennis are raw images, so $\failset$ is defined by any state where the ball has crossed the agent's boundary since this corresponds to the agent losing. It is known that Atari can be solved using a neural network policy~\citep{mnih15}, and this ensures that a policy class parameterized by neural networks is indeed complete.

To ensure that Tennis satisfies the third condition, we need to design an appropriate binary reward function. This is handled by redefining $\failset$ to include any state $s \in \mathcal{S}_{H-1}$ where the ball has not crossed the opposing player's boundary. Then one can simply assign a reward of $1$ to any state in $\mathcal{S}_{H-1} \setminus \failset$, and $0$ to all other states, as required by the Generic Game condition. Hence, playing optimally in this Generic Game framework ensures the ball has moved past the opponent's boundary, corresponding to winning the game.

We now verify that Tennis satisfies the \eff{} condition with a relatively small value of $\C$. In Tennis, after the opposing player hits the ball, the agent must react to the trajectory of the ball and adjust its position accordingly to hit it. If it takes too long to react before it starts adjusting its position, then it will be unable to reach the ball in time. More formally, assume that at timestep $t$ the paddle has not lost the game and the ball is moving towards its boundary. At timestep $t$, the ball may be too close to the boundary, and so the agent will not not have enough time to move its player fast enough in order to reach the ball in time. However, at timestep $t-\C$ the ball is further away from the boundary, so the agent has enough time to move its player appropriately in order to react, reach the ball and hit it back. So at timestep $t-\C$ the agent lies in a safe state in $\mathcal{S}^\star$, since it has enough time to adjust its player and hit the ball back, and hence play optimally. Notably, if we let $\C'$ be the number of timesteps it takes for the ball to traverse from one end of the screen to the other, then $\C \leq \C'$. Hence, when $H$ is large and the agent needs to control its player for many rounds, then $\C$ is a constant independent of $H$.

\begin{figure}[H]
\centering
\setlength{\fboxsep}{0mm}
\setlength{\fboxrule}{0.5mm}
\fcolorbox{black}{white}{\includegraphics[width=0.4\textwidth]{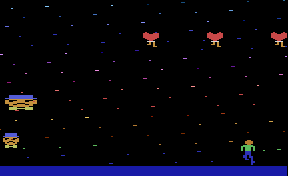}}
\caption{An image of the Atari Journey Escape game. The agent must avoid the enemy objects on screen to avoid receiving a penalty, and must collide with other friendly objects (not depicted) to increase the score.}
\label{fig:journeyescape1}
\end{figure}

\noindent \textbf{Journey Escape.} This game is similar to Skiing, and is depicted in Figure~\ref{fig:journeyescape1}. In this game, there are friendly and enemy objects. These objects come sequentially, and the agent must dodge enemy objects and collide with friendly objects.

The first two conditions of Generic Games are easy to verify. Note that the states in Journey Escape are raw images, so $\failset$ is defined by any state where the agent has collided with an enemy object or missed a friendly object. It is known that Atari can be solved using a neural network policy~\citep{mnih15}, and this ensures that a policy class parameterized by neural networks is indeed complete.

To ensure that Journey Escape satisfies the third condition, we need to design an appropriate binary reward function. This is done by ensuring that $\failset$ includes any state where the agent has collided with an enemy object or missed a friendly object, as described above. Then one can simply assign a reward of $1$ to any state in $\mathcal{S}_{H-1} \setminus \failset$, and $0$ to all other states, as required by the Generic Game condition. Hence, playing optimally in this Generic Game framework ensures the agent has avoided all enemy objects while colliding with all friendly objects, corresponding to winning the game.

We now verify that Journey Escape satisfies the \eff{} condition with a relatively small value of $\C$. As the objects come towards the agent, it must react appropriately to adjust its position depending on whether the oncoming object is friendly or enemy. Let us focus on the enemy object case, since the friendly object case is symmetric. Formally, assume that at timestep $t$ an enemy object is moving towards the agent. At timestep $t$, the object may be too close to the agent, and so the agent will not not have enough time to move away fast enough and get away from the enemy object. However, at timestep $t-\C$ the agent is further away from the enemy object, so the agent has enough time to move away appropriately in order to react. So at timestep $t-\C$ the agent lies in a safe state in $\mathcal{S}^\star$, since it has enough time to adjust its position and hence play optimally. Notably, if we let $\C'$ be the number of timesteps it takes for the agent to traverse from one end of the screen to the other, then $\C \leq \C'$. Hence, when $H$ is large and the agent needs to play for many rounds, then $\C$ is a constant independent of $H$.

\subsection{CoinRun}
\label{app:non_atari}

\begin{figure}[H]
\centering
\setlength{\fboxsep}{0mm}
\setlength{\fboxrule}{0.5mm}
\fcolorbox{black}{white}{\includegraphics[width=0.4\textwidth]{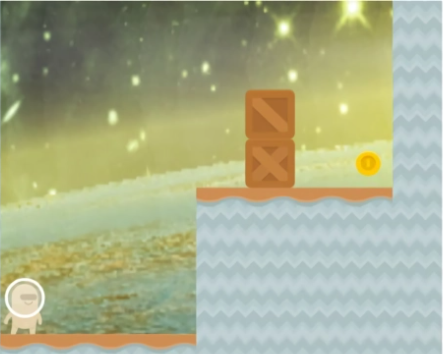}}
\caption{An image of the CoinRun game. The agent must move towards the coin while avoiding obstacles.}
\label{fig:coinrun1}
\end{figure}

CoinRun is a recent RL gaming benchmark~\citep{cobbe19}, and is depicted in Figure~\ref{fig:coinrun1}. In any CoinRun instance, an agent must move right and jump to avoid obstacles, which are sometimes randomly moving, until it arrives at a coin. It receives unit reward if it reaches the coin, and zero reward otherwise. If it collides with an obstacle then the game is over.

The first two conditions of Generic Games are easy to verify. Note that the states in CoinRun are raw images, so $\failset$ is defined by any state where the agent has collided with an obstacle, as well as any state $s \in \mathcal{S}_{H-1}$ where the agent has not already reached the coin. It is known that CoinRun can be solved using a neural network policy~\citep{cobbe19}, and this ensures that a policy class parameterized by neural networks is indeed complete.

Note that the third condition of Generic Games is automatically satisfied by the reward function we described above. This is because the game already has a binary reward function, with unit reward for reaching the coin and completing the game, and zero reward otherwise.

To verify the \eff{} condition, note that the agent must react to obstacles which move towards it. Formally, assume that at timestep $t$ an obstacle is moving towards the agent. At timestep $t$, the obstacle may be too close to the agent, and so the agent will not not have enough time to move away fast enough and get away from the obstacle. However, at timestep $t-\C$ the agent is further away from the obstacle, so the agent has enough time to move away appropriately in order to react. So at timestep $t-\C$ the agent lies in a safe state in $\mathcal{S}^\star$, since it has enough time to adjust its position and hence play optimally. In this game, $\C$ is a small constant, since it only takes a few timesteps for the agent to have enough time to move away from an oncoming obstacle.

\setlength{\parskip}{1em}
\raggedright
\section{Upper Bound Proof}
\label{app:upper_bound}
In this section we will prove Theorem~\ref{thm:main}. First, we shall develop notation and state some helpful lemmas. We then present the proof of Theorem~\ref{thm:main}, and return to complete the proofs of the lemmas.

Recall that Algorithm~\ref{alg:main} iteratively constructs a new $\tvec(t)$ after each timestep in its inner loop. So after $t$ iterations of its inner loop, the algorithm has constructed $\tvec(t)$. For each $t \in \{ 0, 1 \dots H-2 \}$ define the function $L_t: \Theta \to \mathbb{R}$ as follows:
$$
L_t(\theta) = \vert \mathcal{A} \vert^{\C+1} \cdot \mathbb{E}_{\tau \sim \pol{}{}{\tvec(t)}} \left[ \mathbb{I}_{s_{t+1+\C} \in \failset} \prod_{j=0}^{\C} \pol{s_{t+j}}{a_{t+j}}{\theta} \right]
$$
Here, the expectation is over the sampling of trajectories $\tau$ from policy $\pol{}{}{\tvec(t)}$, where $\tau = \left \{ (s_h, a_h)_{h=0}^{H-1} \right \}$. So for each $t \in \{ 0, 1 \dots H-2 \}$, the function $L_t$ is associated with the quantity $\tvec(t)$ that has been constructed by Algorithm~\ref{alg:main}. Observe that the function $\widehat{L}_t$ defined in the inner loop of Algorithm~\ref{alg:main} is the empirical version of $L_t$. Also recall the definition of $\widehat{\theta}_t$ from Eq.~\eqref{eq:emp_loss_minimizer}.

We now define the notation $\mathbb{P}_{\theta}(E ~ | ~ X)$ to denote the probability of event $E$ occurring when using policy $\pol{}{}{\theta}$, conditioned on executing $\pol{}{}{\theta}$ from the state $X$. We now state a key lemma, which is essential to our proof of Theorem~\ref{thm:main}.

\begin{lemma}
\label{lem:recursion}
For any $(\mathcal{M}, \pi(\Theta))$ satisfying the \eff{} condition, and $\tvec(t)$ constructed by Algorithm~\ref{alg:main} for any $t \in \{ 0, 1 \dots H-2 \}$ when given sample size $n$, assume that $\Prob_{\tvec(t)}(s_t \in \mathcal{S}^\star ~ | ~ s_{0}) \geq 1 - \alpha$ for some $\alpha \in (0,1)$. Then the event
$$
\mathbb{P}_{\tvec(t+1)}(s_{t + 1} \notin \mathcal{S}^\star ~ | ~ s_{0}) \leq 2 \vert \mathcal{A} \vert^{\C+1} \sqrt{\frac{\log(2/ \delta')}{n}} + \alpha
$$
holds with probability at least $1 - \delta' \left( 1 + 16 \sqrt{\frac{n}{\log(2/\delta')}} \C \plip \tbound \right)^{\tdim}$.
\end{lemma}

Simply put, the lemma shows the following. Assume that up till timestep $t$, Algorithm~\ref{alg:main} has computed a policy which arrives at a state in $\mathcal{S}^\star$ with high probability. Then at timestep $t+1$ it will compute a policy which arrives at a state in $\mathcal{S}^\star$ with only slightly worse probability. We shall return to prove this lemma in Appendix~\ref{app:proof_lemma5_recursion}. Let us now prove Theorem~\ref{thm:main}.

\subsection{Proof Of Theorem~\ref{thm:main}}
\label{app:proof_thm:main}
Note that by definition of a Generic Game, $\V{s_0}{\pol{}{}{\theta^\star}} = 1$. Furthermore we have 
\begin{align*}
    \V{s_0}{\pol{}{}{\tvec}} &= \Prob_{\tvec} (s_{H-1} \in \mathcal{S}_{H-1} - \failset ~ | ~ s_{0}) \\
    &\geq \Prob_{\tvec} (s_{H-1} \in \mathcal{S}^\star ~ | ~ s_{0})
\end{align*}
where the equality is by the definition of the binary reward function in Generic Games, and the inequality is since the MDP is partitioned into disjoint levels and additionally because $\mathcal{S}^\star \subseteq \mathcal{S} - \failset$. It is hence to sufficient to show that Algorithm~\ref{alg:main} returns $\tvec \equiv \tvec(H-1)$ satisfying
\begin{equation}
\label{eqn:thm_to_show}
\Prob_{\tvec(H-1)} (s_{H-1} \in \mathcal{S}^\star ~ | ~ s_{0}) \geq 1 - \epsilon
\end{equation}
with probability at least $1 - \delta$. We shall devote the remainder of the proof to this.

Let $\delta'$ be some real number in the interval $(0,1)$, whose precise value we will specify later. For each $t \in [H]$, let us define $\mathcal{E}_t$ to be the event that Algorithm~\ref{alg:main} constructs $\tvec(t)$ satisfying
\begin{equation}
\label{eq:thm_induct_event}
\mathbb{P}_{\tvec(t)}(s_{t} \notin \mathcal{S}^\star ~ | ~ s_{0}) \leq t \cdot 2 \vert \mathcal{A} \vert^{\C+1} \sqrt{\frac{\log(2/ \delta')}{n}}.    
\end{equation}
Let $\mathbb{P}_n$ denote the randomness of Algorithm~\ref{alg:main}, which manifests due to sampling of trajectories at each timestep of the inner loop of the algorithm. We claim that
\begin{equation}
\label{eq:thm_induct}
\mathbb{P}_n( \cap_{j \leq t} \mathcal{E}_j) \geq 1 - t \cdot \delta' \left( 1 + 16 \sqrt{\frac{n}{\log(2/\delta')}} \C \plip \tbound \right)^{\tdim}
\end{equation}
for each $t \in [H]$. We will prove this by strong induction, repeatedly using Lemma~\ref{lem:recursion} and union bounding to obtain the desired estimate.

For the base case at timestep $t=0$, notice we trivially have $\Prob_{\tvec(0)}(s_0 \in \mathcal{S}^\star ~ | ~ s_{0}) = 1$, since $s_0 \in \mathcal{S}^\star$ by definition of $\theta^\star$. This implies Eq.~\eqref{eq:thm_induct_event}. In particular, we have $\Prob_{n}(\mathcal{E}_0) = 1$, verifying Eq.~\eqref{eq:thm_induct} when $t=0$.

Now for the inductive step, assume that for some $t$ we have $\mathbb{P}_n\left(\bigcap_{j \leq t} \mathcal{E}_j\right) \geq 1 - t \cdot \delta' \left( 1 + 16 \sqrt{\frac{n}{\log(2/\delta')}} \C \plip \tbound \right)^{\tdim}$. Then by conditioning on $\bigcap\limits_{j \leq t} \mathcal{E}_j$ and applying Lemma~\ref{lem:recursion}, we obtain that the event
\begin{align*}
    \mathbb{P}_{\tvec(t+1)}(s_{t + 1} \in \mathcal{S}^\star ~ | ~ s_{0}) &\geq 1 - 2 \vert \mathcal{A} \vert^{\C+1} \sqrt{\frac{\log(2/ \delta')}{n}} - t \cdot 2 \vert \mathcal{A} \vert^{\C+1} \sqrt{\frac{\log(2/ \delta')}{n}} \\
    &= 1 - (t+1) \cdot 2 \vert \mathcal{A} \vert^{\C+1} \sqrt{\frac{\log(2/ \delta')}{n}}
\end{align*}
holds with probability at least $1 - \delta' \left( 1 + 16 \sqrt{\frac{n}{\log(2/\delta')}} \C \plip \tbound \right)^{\tdim}$. Note that the above equation exactly matches Eq.~\eqref{eq:thm_induct_event}, so conditioned on $\bigcap\limits_{j \leq t} \mathcal{E}_j$ we have shown $\Prob_n(\mathcal{E}_{t+1}) \geq 1 - \delta' \left( 1 + 16 \sqrt{\frac{n}{\log(2/\delta')}} \C \plip \tbound \right)^{\tdim}$. Applying a union bound, we have shown that
$$
\Prob_n\left(\bigcap\limits_{j \leq t+1} \mathcal{E}_{j}\right) \geq 1 - (t+1) \cdot \delta' \left( 1 + 16 \sqrt{\frac{n}{\log(2/\delta')}} \C \plip \tbound \right)^{\tdim},
$$
which thus verifies Eq.~\eqref{eq:thm_induct} and hence the inductive step. We have therefore shown that Algorithm~\ref{alg:main} constructs $\tvec(H-1)$ satisfying
$$
\Prob_{\tvec(H-1)} (s_{H-1} \in \mathcal{S}^\star ~ | ~ s_{0}) \geq 1 - 2 H \vert \mathcal{A} \vert^{\C+1} \sqrt{\frac{\log(2/ \delta')}{n}} 
$$
with probability at least $1 - H \delta' \left( 1 + 16 \sqrt{\frac{n}{\log(2/\delta')}} \C \plip \tbound \right)^{\tdim}$. To show that the above equation exactly matches Eq.~\eqref{eqn:thm_to_show}, we need only check that our choice of $n$ yields the desired values of $\epsilon$ and $\delta$.

To obtain our choice of $n$, we first set \(2H \vert \mathcal{A} \vert^{\C+1} \sqrt{\frac{\log(2/ \delta')}{n}} = \epsilon\).
This yields
\begin{equation}
\label{eq:thm_n_helper1}
    n = \frac{4H^{2} \vert \mathcal{A} \vert^{2\C + 2}}{\epsilon^{2}} \log\left(\frac{2}{\delta'}\right) \iff \sqrt{\frac{n}{\log(2 / \delta')}} = \frac{2H \vert \mathcal{A} \vert^{\C + 1}}{\epsilon}.
\end{equation}
Next, we make the substitution
\begin{equation*}
    \delta = H\delta'\left(1 + 16\sqrt{\frac{n}{\log(2 / \delta')}}\C\plip \tbound\right)^{\tdim},
\end{equation*}
which results in
\begin{align*}
    \delta = H\delta'\left(1 + \frac{32H \vert \mathcal{A} \vert^{\C + 1} }{\epsilon} \C \plip \tbound\right)^{\tdim},
\end{align*}
implying that
$$
    \delta' = \delta \left(H \left(1 + \frac{32 H \vert \mathcal{A} \vert^{\C + 1}}{\epsilon} \C \plip \tbound\right)^{\tdim}\right)^{-1}.
$$
Substituting the above expression for $\delta'$ into the first equivalence of Eq.~\eqref{eq:thm_n_helper1}, we finally get
\begin{equation*}
    n = \frac{4H^{2} \vert \mathcal{A} \vert^{2\C + 2}}{\epsilon^{2}} \left(\log\left(\frac{2H}{\delta}\right) + \tdim\log\left(1 + \frac{32 H \vert \mathcal{A} \vert^{\C + 1} \C \plip \tbound}{\epsilon} \right)\right),
\end{equation*}
and this completes the proof.

\subsection{Proof Of Lemma~\ref{lem:recursion}}
\label{app:proof_lemma5_recursion}
To facilitate our proof, we first state the following useful lemma. Recall the definition of $\widehat{\theta}_t$ from Eq.~\eqref{eq:emp_loss_minimizer}.

\begin{lemma}
\label{lem:main_empirical_bound}
For any $(\mathcal{M}, \pi(\Theta))$ satisfying the \eff{} condition, and $\tvec(t)$ constructed by Algorithm~\ref{alg:main} for any $t \in \{ 0, 1 \dots H-2 \}$ when given sample size $n$, assume that $\Prob_{\tvec(t)}(s_t \in \mathcal{S}^\star ~ | ~ s_{0}) \geq 1 - \alpha$ for some $\alpha \in (0,1)$. Then the event
$$
\mathbb{E}_{s_t \sim \pol{}{}{\tvec(t)}}\left[\mathbb{P}_{\widehat{\theta}_t}(s_{t + \C + 1} \in \failset ~ | ~ s_{t})\right] \leq 2 \vert \mathcal{A} \vert^{\C+1} \sqrt{\frac{\log(2/ \delta')}{n}} + \alpha
$$
holds with probability at least $1 - \delta' \left( 1 + 16 \sqrt{\frac{n}{\log(2/\delta')}} \C \plip \tbound \right)^{\tdim}$.
\end{lemma}

We will return to prove Lemma~\ref{lem:main_empirical_bound} in Appendix~\ref{app:proof_lem:main_empirical_bound}. For now, let us return to the proof of Lemma~\ref{lem:recursion}. By the result of Lemma~\ref{lem:main_empirical_bound}, the event
\begin{equation*}
\mathbb{E}_{s_t \sim \pol{}{}{\tvec(t)}}\left[\mathbb{P}_{\widehat{\theta}_t}(s_{t + \C + 1} \in \failset ~ | ~ s_{t})\right] \leq 2 \vert \mathcal{A} \vert^{\C+1} \sqrt{\frac{\log(2/ \delta')}{n}} + \alpha
\end{equation*}
holds with probability at least $1 - \delta' \left( 1 + 16 \sqrt{\frac{n}{\log(2/\delta')}} \C \plip \tbound \right)^{\tdim}$. Let us denote this event as $\mathcal{E}$. Then on this event, we have
\begin{align*}
    \mathbb{E}_{s_t \sim \pol{}{}{\tvec(t)}}\left[\mathbb{P}_{\widehat{\theta}_t}(s_{t + 1} \notin \mathcal{S}^\star ~ | ~ s_{t})\right] &= \mathbb{E}_{s_t \sim \pol{}{}{\tvec(t)}}\left[\mathbb{P}_{\widehat{\theta}_t}(s_{t + 1} \notin \mathcal{S}^\star \wedge s_{t + \C + 1} \in \failset ~ | ~ s_{t}) \right.\\
    &\qquad \left.+~\mathbb{P}_{\widehat{\theta}_t}(s_{t + 1} \notin \mathcal{S}^\star \wedge s_{t + \C + 1} \notin \failset ~ | ~ s_{t}) \right] \\
    &\overset{(i)}\leq \mathbb{E}_{s_t \sim \pol{}{}{\tvec(t)}}\left[\mathbb{P}_{\widehat{\theta}_t}(s_{t + \C + 1} \in \failset ~ | ~ s_{t}) + \mathbb{P}_{\widehat{\theta}_t}(s_{t + 1} \notin \mathcal{S}^\star \wedge s_{t + \C + 1} \notin \failset ~ | ~ s_{t}) \right] \\
    &\overset{(ii)}\leq 2 \vert \mathcal{A} \vert^{\C+1} \sqrt{\frac{\log(2/ \delta')}{n}} + \alpha + \mathbb{E}_{s_t \sim \pol{}{}{\tvec(t)}}\left[ \mathbb{P}_{\widehat{\theta}_t}(s_{t + 1} \notin \mathcal{S}^\star \wedge s_{t + \C + 1} \notin \failset ~ | ~ s_{t}) \right] \\
    &\overset{(iii)}= 2 \vert \mathcal{A} \vert^{\C+1} \sqrt{\frac{\log(2/ \delta')}{n}} + \alpha.
\end{align*}
Here, step \((i)\) is trivial as \(\mathbb{P}(A \wedge B) \leq \mathbb{P}(A)\). Step \((ii)\) follows from the definition of $\mathcal{E}$, and finally, step \((iii)\) follows from the \eff{} condition and definition of $\C$.
It remains to note that
$$
\mathbb{E}_{s_t \sim \pol{}{}{\tvec(t)}}\left[\mathbb{P}_{\widehat{\theta}_t}(s_{t + 1} \notin \mathcal{S}^\star ~ | ~ s_{t})\right] = \mathbb{P}_{\tvec(t+1)}(s_{t + 1} \notin \mathcal{S}^\star ~ | ~ s_{0}),
$$
which follows directly from the definition $\tvec(t+1) = [\widehat{\theta}_0, \widehat{\theta}_1 \ldots, \widehat{\theta}_{t-1}, \widehat{\theta}_t, \trand, \ldots, \trand]$ and the Law of Total Expectation.

\subsection{Proof Of Lemma~\ref{lem:main_empirical_bound}}
\label{app:proof_lem:main_empirical_bound}
To facilitate this proof, we first state the following two auxiliary lemmas.

\begin{lemma}
\label{lem:pop_prob}
For any $(\mathcal{M}, \pi(\Theta))$ satisfying the \eff{} condition, and $\tvec(t)$ constructed by Algorithm~\ref{alg:main} for any $t \in \{ 0, 1 \dots H-2 \}$, we have
\begin{equation*}
    L_{t}(\theta) = \mathbb{E}_{s_t \sim \pol{}{}{\tvec(t)}}\left[\mathbb{P}_{\theta}(s_{t + \C + 1} \in \failset ~ | ~ s_{t})\right],
\end{equation*}
where the expectation is taken with respect to the marginal distribution of \(s_{t}\) while sampling trajectories from \(\pol{}{}{\tvec}\).
\end{lemma}

\begin{lemma}
\label{lem:lip_loss}
For any $(\mathcal{M}, \pi(\Theta))$ satisfying the \eff{} condition, and $\tvec(t)$ constructed by Algorithm~\ref{alg:main} for any $t \in \{ 0, 1 \dots H-2 \}$, the functions $L_t$, $\widehat{L}_t$ are each Lipschitz with Lipschitz constant $\vert \mathcal{A} \vert^{\C+1} (\C+1) \plip$.
\end{lemma}

We shall return to prove these lemmas in Appendices~\ref{app:proof_lem:pop_prob} and~\ref{app:proof_lem:lip_loss} respectively. Let us now return to the proof of Lemma~\ref{lem:main_empirical_bound}. Recall from Lemma~\ref{lem:pop_prob} that $\mathbb{E}_{s_t \sim \pol{}{}{\tvec(t)}}\left[\mathbb{P}_{\widehat{\theta}_t}(s_{t + \C + 1} \in \failset ~ | ~ s_{t})\right] = L_t(\widehat{\theta}_t)$. So it is sufficient to show that the event
$$
L_t(\widehat{\theta}_t) \leq 2 \vert \mathcal{A} \vert^{\C+1} \sqrt{\frac{\log(2/ \delta')}{n}} + \alpha
$$
holds with probability at least $1 - \delta' \left( 1 + 16 \sqrt{\frac{n}{\log(2/\delta')}} \C \plip \tbound \right)^{\tdim}$, and we will devote the remainder of the proof to showing this. First, we use the characterization of $L_t$ derived in Lemma~\ref{lem:pop_prob} to show the helpful fact that
\begin{equation}
\begin{aligned}
\label{eq:pop_opt_erm_bound}
    L_t(\theta^\star) &= \mathbb{E}_{s_t \sim \pol{}{}{\tvec(t)}}\left[\mathbb{P}_{\theta^\star}(s_{t + \C + 1} \in \failset ~ | ~ s_{t})\right] \\
    &= \mathbb{E}_{s_t \sim \pol{}{}{\tvec(t)}}\left[\mathbb{P}_{\theta^\star}(s_{t + \C + 1} \in \failset ~ | ~ s_{t}) \mathbb{I}_{s_t \in \mathcal{S}^\star} + \mathbb{P}_{\theta^\star}(s_{t + \C + 1} \in \failset ~ | ~ s_{t}) \mathbb{I}_{s_t \notin \mathcal{S}^\star} \right] \\
    &= \mathbb{E}_{s_t \sim \pol{}{}{\tvec(t)}}\left[\mathbb{P}_{\theta^\star}(s_{t + \C + 1} \in \failset ~ | ~ s_{t}) \mathbb{I}_{s_t \notin \mathcal{S}^\star} \right] \\
    &\leq \alpha,
\end{aligned}
\end{equation}
where the final equality follows from the definition of $\mathcal{S}^\star$ and the inequality follows from the assumption that $\Prob_{\tvec(t)}(s_t \in \mathcal{S}^\star ~ | ~ s_{0}) \geq 1 - \alpha$.

By the Regularity of $\pol{}{}{\Theta}$, we are guaranteed that $\Theta$ is contained in the Euclidean ball of radius $\tbound$. For any $\gamma > 0$, we use $\mathcal{N}(\gamma)$ to denote a minimal $\gamma$-covering of $\Theta$. Recall that $\Theta \subset \mathbb{R}^{\tdim}$. Also recall the standard fact~\citep{vershynin2018high} that $\vert \mathcal{N}(\gamma) \vert \leq \left( 1 + \frac{2\tbound}{\gamma} \right)^{\tdim}$.

Now for any fixed $\theta \in \mathcal{N}(\gamma)$, we know from Hoeffding's inequality that the bound
\begin{align}
\label{eq:empirical_pop_bound}
    \vert \widehat{L}_t(\theta) - L_t(\theta) \vert \leq \frac{\vert \mathcal{A} \vert^{\C+1}}{2} \sqrt{\frac{\log(2/ \delta')}{n}}
\end{align}
holds with probability at least $1 - \delta'$. Hence, applying a union bound, we know that the above bound holds for every $\theta \in \mathcal{N}(\gamma)$ with probability at least $1 - \delta' \vert \mathcal{N}(\gamma) \vert \geq 1 - \delta' \left( 1 + \frac{2\tbound}{\gamma} \right)^{\tdim}$.

For any $\theta \in \Theta$, let $\theta_{\gamma}$ be an element of $\mathcal{N}(\gamma)$ such that $\| \theta - \theta_{\gamma} \|_2 \leq \gamma$.
We now argue that with probability at least $1 - \delta' \left( 1 + \frac{2\tbound}{\gamma} \right)^{\tdim}$, any $\theta \in \Theta$ satisfies the bound
\begin{align*}
    \vert \widehat{L}_t(\theta) - L_t(\theta) \vert &\overset{(i)}\leq \vert \widehat{L}_t(\theta) - \widehat{L}_t(\theta_{\gamma}) \vert + \vert \widehat{L}_t(\theta_{\gamma}) - L_t(\theta_{\gamma}) \vert + \vert L_t(\theta_{\gamma}) - L_t(\theta) \vert \\
    &\overset{(ii)}\leq 2 \vert \mathcal{A} \vert^{\C+1} (\C+1) \plip \gamma + \vert \widehat{L}_t(\theta_{\gamma}) - L_t(\theta_{\gamma}) \vert \\
    &\overset{(iii)}\leq 2 \vert \mathcal{A} \vert^{\C+1} (\C+1) \plip \gamma + \frac{\vert \mathcal{A} \vert^{\C+1}}{2} \sqrt{\frac{\log(2/ \delta')}{n}},
\end{align*}
where step \((i)\) follows from the triangle inequality, step \((ii)\) is due to the Lipschitz property of $L_t, \widehat{L}_t$ we derived in Lemma~\ref{lem:lip_loss}, and step \((iii)\) is due to Eq.~\eqref{eq:empirical_pop_bound}. Now set $\gamma = \frac{1}{4(\C+1)\plip} \sqrt{\frac{\log(2/ \delta')}{n}}$. Then the bound
\begin{align}
\label{eq:erm_bound_helper1}
\vert \widehat{L}_t(\theta) - L_t(\theta) \vert \leq \vert \mathcal{A} \vert^{\C+1} \sqrt{\frac{\log(2/ \delta')}{n}}
\end{align}
holds for uniformly for each $\theta \in \Theta$ with probability at least $1 - \delta' \left( 1 + 16 \sqrt{\frac{n}{\log(2/\delta')}} \C \plip \tbound \right)^{\tdim}$. Let $\mathcal{E}$ denote the event that Eq.~\eqref{eq:erm_bound_helper1} holds uniformly for each $\theta \in \Theta$.

We now use this uniform bound to control the quantity $L_t(\widehat{\theta}_t)$. Concretely, on the event $\mathcal{E}$ we have that
\begin{align*}
    L_t(\widehat{\theta}_t) &= L_t(\widehat{\theta}_t) - \widehat{L}_t(\widehat{\theta}_t) + \widehat{L}_t(\widehat{\theta}_t) - \widehat{L}_t(\theta^\star) + \widehat{L}_t(\theta^\star) \\
    &\overset{(iv)}\leq L_t(\widehat{\theta}_t) - \widehat{L}_t(\widehat{\theta}_t) + \widehat{L}_t(\theta^\star) \\
    &\overset{(v)}\leq \vert \mathcal{A} \vert^{\C+1} \sqrt{\frac{\log(2/ \delta')}{n}} + \widehat{L}_t(\theta^\star),
\end{align*}
where step \((iv)\) follows from the definition $\widehat{\theta}_t \in \argmin_{\theta} \widehat{L}_t(\theta)$, and step \((v)\) follows from Eq.~\eqref{eq:erm_bound_helper1}. To obtain control on $L_t(\widehat{\theta}_t)$, it remains to bound $\widehat{L}_t(\theta^\star)$. Simply observe that on the event $\mathcal{E}$ we have
\begin{align*}
    \widehat{L}_t(\theta^\star) &= \widehat{L}_t(\theta^\star) - L_t(\theta^\star) + L_t(\theta^\star) \\
    &\overset{(vi)}\leq \vert \mathcal{A} \vert^{\C+1} \sqrt{\frac{\log(2/ \delta')}{n}} + L_t(\theta^\star) \\
    &\overset{(vii)}\leq \vert \mathcal{A} \vert^{\C+1} \sqrt{\frac{\log(2/ \delta')}{n}} + \alpha.
\end{align*}
Steps \((vi)\) and \((vii)\) follow from Eq.~\eqref{eq:erm_bound_helper1} and Eq.~\eqref{eq:pop_opt_erm_bound} respectively.
Putting the previous two equations together and recalling the definition of $\mathcal{E}$, we have demonstrated that the bound
$$
L_t(\widehat{\theta}_t) \leq 2 \vert \mathcal{A} \vert^{\C+1} \sqrt{\frac{\log(2/ \delta')}{n}} + \alpha
$$
holds with probability at least $1 - \delta' \left( 1 + 16 \sqrt{\frac{n}{\log(2/\delta')}} \C \plip \tbound \right)^{\tdim}$. As argued earlier, this is sufficient to complete the proof.

\subsection{Proof Of Lemma~\ref{lem:pop_prob}}
\label{app:proof_lem:pop_prob}
Recall that notation $\tau = \left \{ (s_h, a_h)_{h=0}^{H-1} \right \}$. Also recall from Algorithm~\ref{alg:main} that at timestep $t$ onwards, $\pol{}{}{\tvec(t)}$ executes $\trand$, implying it selects actions uniformly at random regardless of the state. This fact allows us to decompose $L_t(\theta)$ as follows
\begin{align*}
    L_t(\theta) &= \vert \mathcal{A} \vert^{\C+1} \cdot \mathbb{E}_{\tau \sim \pol{}{}{\tvec(t)}} \left[ \mathbb{I}_{s_{t+1+\C} \in \failset} \prod_{j=0}^{\C} \pol{s_{t+j}}{a_{t+j}}{\theta} \right] \\
    &= \vert \mathcal{A} \vert^{\C+1} \cdot \mathbb{E}_{s_t \sim \pol{}{}{\tvec(t)}} \left[ \mathbb{E}_{\pol{}{}{\trand}} \left( \mathbb{I}_{s_{t+1+\C} \in \failset} \left. \prod_{j=0}^{\C} \pol{s_{t+j}}{a_{t+j}}{\theta} ~ \right\vert ~ s_t \right) \right] \\
    &= \mathbb{E}_{s_t \sim \pol{}{}{\tvec(t)}} \left[ \mathbb{E}_{\pol{}{}{\theta}} \left( \left. \mathbb{I}_{s_{t+1+\C} \in \failset} ~ \right\vert ~ s_t \right) \right] \\
    &= \mathbb{E}_{s_t \sim \pol{}{}{\tvec(t)}}\left[\mathbb{P}_{\theta}(s_{t + \C + 1} \in \failset ~ | ~ s_{t})\right].
\end{align*}
This completes the proof.

\subsection{Proof Of Lemma~\ref{lem:lip_loss}}
\label{app:proof_lem:lip_loss}
We first note that the product of \(m \geq 2\) functions \(\{f_{i}\}_{i=1}^{m}\) which are bounded by \(1\) and Lipschitz continuous with constant \(L\) is also Lipschitz continuous with constant \(mL\).
This can be proved by induction.
Consider the base case when \(m = 2\).
For any \(x, y\) in the domain, we have
\begin{align*}
    |f_{1}(x)f_{2}(x) - f_{1}(y)f_{2}(y)| &= |f_{1}(x)f_{2}(x) - f_{1}(x)f_{2}(y) + f_{1}(x)f_{2}(y) - f_{1}(y)f_{2}(y)| \\
    &\overset{(i)}\leq |f_{1}(x)|~|f_{2}(x) - f_{2}(y)| + |f_{2}(y)|~|f_{1}(x) - f_{2}(y)| \\
    &\overset{(ii)}\leq |f_{2}(x) - f_{2}(y)| + |f_{1}(x) - f_{1}(y)| \\
    &\overset{(iii)}\leq 2L\|x - y\|_{2},
\end{align*}
where in steps \((i)\), \((ii)\) and \((iii)\) we have used the triangle inequality, the fact that \(f_{1}\), \(f_{2}\) are bounded by \(1\) and the Lipschitz continuity of \(f_{1}\),  \(f_{2}\) respectively.

Now assume that \(g_{[k]} = f_{1} \ldots f_{k}\) is Lipschitz continuous with constant \(kL\) for some \(k \geq 2\).
Following the same steps from above, we see that \(g_{[k]}f_{k+1}\) is \((k + 1)L\)-Lipschitz.
This completes the proof of the fact.
%Therefore, we have shown by induction that the product of \(n\) functions bounded by \(1\) and which are \(L\)-Lipschitz is \(nL\)-Lipschitz.

We now prove the statement of the lemma.
Let $\theta, \theta'$ be two distinct policy parameters.
For any \(h \in [H]\), we have
\begingroup
\allowdisplaybreaks
\begin{align*}
\left| L_h(\theta) - L_h(\theta') \right|
&= \vert \mathcal{A} \vert^{\C+1} \cdot \left| \mathbb{E}_{\tau \sim \pol{}{}{\tvec}} \left[ \mathbb{I}_{s_{t+1+C} \in \failset} \prod_{j=0}^{\C} \pol{s_{t+j}}{a_{t+j}}{\theta} \right] - \mathbb{E}_{\tau \sim \pol{}{}{\tvec}} \left[ \mathbb{I}_{s_{t+1+C} \in \failset} \prod_{j=0}^{\C} \pol{s_{t+j}}{a_{t+j}}{\theta'} \right]\right| \\
&= \vert \mathcal{A} \vert^{\C+1} \cdot \left| \mathbb{E}_{\tau \sim \pol{}{}{\tvec}} \left[ \mathbb{I}_{s_{t+1+C} \in \failset} \left(\prod_{j=0}^{\C} \pol{s_{t+j}}{a_{t+j}}{\theta} - \prod_{j=0}^{\C} \pol{s_{t+j}}{a_{t+j}}{\theta'}\right) \right]\right| \\
&\overset{(iv)}\leq \vert \mathcal{A} \vert^{\C+1} \cdot \mathbb{E}_{\tau \sim \pol{}{}{\tvec}} \left[ \mathbb{I}_{s_{t+1+C} \in \failset} \left(\left| \prod_{j=0}^{\C} \pol{s_{t+j}}{a_{t+j}}{\theta} - \prod_{j=0}^{\C} \pol{s_{t+j}}{a_{t+j}}{\theta'} \right| \right) \right] \\
&\overset{(v)}\leq \vert \mathcal{A} \vert^{\C+1} \cdot \mathbb{E}_{\tau \sim \pol{}{}{\tvec}} \left[\mathbb{I}_{s_{t+1+C} \in \failset} \cdot \left((C + 1)\plip \|\theta - \theta'\|_{2}\right) \right] \\
&\leq |\mathcal{A}|^{C+1}(C+1)\plip\| \theta - \theta'\|_{2}.
\end{align*}
\endgroup
Step \((iv)\) is due to Jensen's inequality, and step \((v)\) is due the Lipschitz continuity of a product of Lipschitz continuous functions bounded by \(1\) which was shown earlier.
Since the functions are policy probabilities, they are bounded by \(1\), and are also \(\plip\)-Lipschitz due to the Regularity of \(\pol{}{}{\Theta}\).

Analogously, we can show the Lipschitz continuity of \(\widehat{L}_{h}\) for any \(h \in \{0, 1 \ldots, H-2\}\), by replacing the expectation with the empirical average over trajectory samples, and this completes the proof of the Lemma.
%\begin{align*}
 %   \mathbb{E}_{s_t \sim \pol{}{}{\tvec(t)}}\left[\mathbb{P}_{\widehat{\theta}_t}(s_{t + 1} \notin \mathcal{S}^* ~ | ~ s_{t})\right] &= \sum_{s_t} \Prob_{\tvec(t)}(s_t) \mathbb{P}_{\widehat{\theta}_t}(s_{t + 1} \notin \mathcal{S}^* ~ | ~ s_{t}) \\
  %  &= \sum_{s_t} \Prob_{\tvec(t+1)}(s_t) \mathbb{P}_{\tvec(t+1)}(s_{t + 1} \notin \mathcal{S}^* ~ | ~ s_{t}) \\
   % &= \sum_{s_t} \mathbb{P}_{\tvec(t+1)}(s_{t+1} \notin \mathcal{S}^* \cap s_t) \\
%    &= \mathbb{P}_{\tvec(t+1)}(s_{t+1} \notin \mathcal{S}^*)
%\end{align*}
\section{Lower Bound Proof Sketch}
\label{app:lower_bound}
As discussed earlier, this result follows almost directly from the results of Du et al.~\citep{du20lowerbound}, and so we only sketch the proof. First we note it is well known that softmax linear policies are Lipschitz~\citep{agarwal20}. In our proof, we use $\Theta$ as the scaled unit ball in $\mathbb{R}^\tdim$, where the scaling factor is polynomial in $H, \mathcal{A}$. Hence the policy class $\pol{}{}{\Theta}$ is indeed Regular. For the construction of $\mathcal{M}$, we use the same construction that was given in the proof of Theorem 4.1 in~\citep{du20lowerbound}. Recall this construction is defined by a horizon $H$ MDP $\mathcal{M}$ whose states, actions and transitions are defined by a binary tree with $H$ levels. There is a single state on the final level with unit reward, and all the other states in the tree have zero reward. To cast this construction in our Generic Game framework, we only need to make a slight modification. For each state on the penultimate level whose child does not have reward, modify it transitions so that taking any action from here deterministically exits the MDP. Then discard each of the now unreachable states on the final level, so the final level only contains a state with unit reward. The set $\failset$ is precisely defined by the states on the penultimate level of the tree from where taking any action exits the MDP. The binary rewards property is true by definition. And the complete policy class property is true directly by the proof of~\citep{du20lowerbound}. Note here, that since we have a bounded $\Theta$ and are using a softmax linear policy class $\pol{}{}{\Theta}$, the $\theta^*$ does not lead to $\mathcal{S}_{H-1} - \failset$ almost surely. However, since $\tbound$ is polynomial in $H, \mathcal{A}$, using $\theta^*$ will lead to $\mathcal{S}_{H-1} - \failset$ with probability exponentially large in $H$. Our main Theorem~\ref{thm:main} easily handles this.

It remains to show the existence of $f: \mathbb{R}^{\sdim} \to \mathbb{R}$ where $f$ is a linear combination of two neurons, and $V^\star_{\mathcal{M}}(s) = f(s)$ for each $s \in \mathcal{S}$. Again, this follows almost immediately from the proof provided by~\citep{du20lowerbound}. Recall that for $\sdim$ sufficiently large, their proof demonstrates the existence of $\theta^{\star \star}$ such that $s^T \theta^{\star \star} = 1$ if $V^\star_{\mathcal{M}}(s) = 1$ and $s^T \theta^{\star \star} \leq 0.25$ if $V^\star_{\mathcal{M}}(s) = 0$. It remains to observe that the function $f$ defined as
$$
f(x) = \relu(2 x^T \theta^{\star \star} - 1) - \relu(-2 x^T \theta^{\star \star} - 1)
$$
exactly satisfies the claim.

\bibliographystyle{alpha}
\bibliography{epw_main}

\end{document}